\documentclass[conference]{IEEEtran}
\IEEEoverridecommandlockouts
\usepackage{cite}
\usepackage{amsmath,amssymb,amsfonts}
\usepackage{algorithmic}
\usepackage{graphicx}
\usepackage{textcomp}
\usepackage{xcolor}
\def\BibTeX{{\rm B\kern-.05em{\sc i\kern-.025em b}\kern-.08em
    T\kern-.1667em\lower.7ex\hbox{E}\kern-.125emX}}

\usepackage{cite}       
\usepackage{url}
\usepackage{booktabs}
\usepackage{tabularx}
\usepackage[vlined,ruled]{algorithm2e}
\usepackage{makecell} 
\usepackage{pgfplots}
\usepackage{etoolbox}
\usepackage[most]{tcolorbox}
\usepackage{hyperref}

\begin{document}

\title{Adaptive Federated Distillation for Multi-Domain Non-IID Textual Data\\
}

\author{

\IEEEauthorblockN{Jiahao Xiao\textsuperscript{$\dag$}, Jiangming Liu\textsuperscript{$\dag \ddag *$}}

\IEEEauthorblockA{
\textsuperscript{$\dag$}School of Information Science and Engineering, Yunnan University, China\\
\textsuperscript{$\ddag$}Yunnan Key Laboratory of Intelligent Systems and Computing, China\\
\textsuperscript{*}Corresponding Author: jiangmingliu@ynu.edu.cn}
}


\maketitle

\begin{abstract}
The widespread success of pre-trained language models has established a new training paradigm, where a global PLM is fine-tuned using task-specific data from local clients. The local data are highly different from each other and can not capture the global distribution of the whole data in real world. To address the challenges of non-IID data in real environments, privacy-preserving federated distillation has been proposed and highly investigated. However, previous experimental non-IID scenarios are primarily identified with the label (output) diversity, without considering the diversity of language domains (input) that is crucial in natural language processing. In this paper, we introduce a comprehensive set of multi-domain non-IID scenarios and propose a unified benchmarking framework that includes diverse data. The benchmark can be used to evaluate the federated learning framework in a real environment. To this end, we propose an \textbf{Ada}ptive \textbf{F}ederated \textbf{D}istillation (\textbf{AdaFD}) framework designed to address multi-domain non-IID challenges in both homogeneous and heterogeneous settings. Experimental results demonstrate that our models capture the diversity of local clients and achieve better performance compared to the existing works. The code for this paper is available at: \href{https://github.com/jiahaoxiao1228/AdaFD}{https://github.com/jiahaoxiao1228/AdaFD}.
\end{abstract}


\section{Introduction}
The widespread success of pre-trained language models (PLMs) in various natural language processing tasks has ushered in a new training paradigm where a global PLM is typically fine-tuned with task-specific data contained by local clients. The decentralized training prevents the clients from explicitly exchanging their local data due to concerns of data security~\cite{chik2013singapore,protection2018general}. 
Federated learning~\cite{mcmahan2017communication} is introduced as a decentralized training approach that prioritizes privacy. 
In this framework, a collective of clients is coordinated by a central server to train a global model by aggregating the local models. 
This approach allows for the effective utilization of private data from clients through the exchange of model parameters, resulting in a unified model that achieves significantly better performance compared to models trained individually.


Previous research on federated learning has developed various frameworks for natural language processing and introduced benchmarking frameworks~\cite{tian2022fedbert,dong2022collaborating,zhang2022federated,lin2022fednlp,abadeer2022flightner,ma-etal-2023-fedid}. Traditional federated learning approaches, however, are constrained by the requirement for identical model architectures across both the server and clients, which prevents clients from independently designing models tailored to their diverse system resources and non-IID (non-Independent Identically Distributed) data. Additionally, the frequent exchange of model parameters incurs significant communication costs. These limitations substantially restrict the applicability and scalability of federated learning for PLMs. To alleviate the problem, federated distillation is proposed by introducing an unlabeled public proxy data, allowing federated learning among heterogeneous models with flexible training~\cite{jeong2018communication,li2019fedmd,chang2019cronus,itahara2021distillation,hu2021mhat,gong2022preserving,ma-etal-2023-fedid} on non-IID data.

However, in natural language processing, most of the federated learning and distillation models with non-IID data focus on the diversity of labels (output) randomly partitioned from a Dirichlet distribution, without considering the diversity of inputs. Different from the images, languages differ across domains due to the vocabulary and communication goals~\cite{fishman2020speaks}. To complete the non-IID problems in federated learning for natural language processing, we investigate the non-IID problem by considering the diversity of inputs and outputs. According to the diversity of inputs and outputs, the federated learning environments can be categorized as follows: IID, multi-domain, label diverse (previous non-IID) and non-IID \footnote{Non-IID in this paper is also identified as multi-domain non-IID.}. 
Federated learning in such a non-IID environment presents more challenges due to the heterogeneity in both label and vocabulary distributions.

To address the above non-IID challenge, existing works have proposed various weight allocation strategies for aggregating logits from clients. FedAvg introduces an equal weighting strategy, while DS-FL\cite{itahara2021distillation}, MHAT\cite{hu2021mhat}, and FedKD\cite{gong2022preserving} allocate weights based on the size of training data. A key limitation is that it can not accurately and dynamically reflect the model's fit to the global data distribution. To address this issue, the \textbf{Ada}ptive \textbf{F}ederated \textbf{D}istillation (AdaFD) framework based on heterogeneous pretrained language models is proposed, which consists of two key components: (1) an adaptive ensemble strategy, comprising three weighting methods based on training loss---Reciprocal Normalized Weight Calculation (RNWC), Exponential Normalized Weight Calculation (ENWC), and Large Language Models Weight Calculation (LLMWC)---where models better aligned with the global data distribution are assigned higher weights, and (2) an adaptive distillation strategy. Our empirical experiments demonstrate that AdaFD outperforms existing methods in both homogeneous and heterogeneous federated settings. The main contributions of this paper are as follows:
\begin{itemize}
    \item We comprehensively introduce non-IID scenarios in natural language processing by considering linguistic diversity together with label diversity.
    \item We present a unified benchmarking framework featuring diverse data distributions across multiple domains, aimed at advancing research on non-IID challenges in natural language processing.
    \item We introduce an adaptive federated distillation framework specifically designed for non-IID problems in both homogeneous and heterogeneous environments.
\end{itemize}

\section{Related Work}

\subsection{Federated Learning and Distillation}

Federated learning has garnered significant interest in natural language processing tasks \cite{ge2020fedner,sui2020feded,tian2022fedbert,wu2023faster,huang2023rethinking,liu2024vertical,yazdinejad2024robust} due to its potential for collaborative training on distributed data while maintaining data privacy \cite{mcmahan2017communication,liu2021federated}. These methods optimize the central models by facilitating collaboration among clients through the exchange of model parameters. However, this parameter exchange poses privacy risks \cite{carlini2021extracting} and restricts the flexibility of client-side model architectures. To mitigate these issues, federated distillation has been proposed, leveraging a public data \cite{lin2020ensemble,hu2021mhat,tan2022fedproto,gong2022preserving,zhang2023distill,li2023feddkd,wang2023dafkd,yang2024fedfed,zhou2024federated,han2024fedal}. In these approaches, client predictions on the public data are aggregated at a central server following the teacher-student distillation framework, enabling knowledge transfer while addressing privacy concerns.

\subsection{Non-IID in Federated Learning}
A key challenge in federated learning is that the data contained in clients is non-IID, i.e., the distribution of each local data is highly
different from the global distribution~\cite{kairouz2021advances,li2021survey}. In federated learning and distillation, the local models are updated to a local optimized point, which can be far from the global optimization. Several models are proposed to address the challenge~\cite{li2021model,wang2021addressing,zhou2023efficient,azam2023federated,yang2023personalized,arafeh2023data,lu2024federated,li2024feature}. However, the current non-IID experimental settings only focus on the different distributions on labels of local training data across clients without considering the inputs, which is significantly important in natural language processing. Instead, we identify the non-IID in natural language processing tasks by discussing both non-IID on labels and non-IID on languages to benchmark the non-IID in natural language processing, contributing to the federated learning community.

\section{Preliminaries}
\subsection{Problem Definition}
In the context of federated distillation with $K$ clients and one server, each client possesses its own labeled private data ${D^k}=\{(x_i^k,y_i^k)|i=1,2,...,|D^k|\}$, $ k\in\{1,2,...,K\}$, isolated from others, while the server only shares access to an unlabeled public data $D^0=\{x_i^0|i=1,2,...,|D^0|\}$ with the clients without accessing to any labeled private data, where $x_i$ and $y_i$ are the $i$th instance and its label, respectively. The objective is obtaining the final central model using decentralized training method to assimilate the knowledge of each client.

\subsection{Federated Distillation}
In the general framework of federated distillation, within a limited number of communication rounds $T$, each client locally initializes and optimizes its own parameters $\theta^k~(k\in\{1,2,...,K\})$ on its own private data ${D^k}$. The server optimizes the central model by aggregating predictions on public data provided by clients, and then returns the global prediction on public data $D^0$ to each client. The clients further update their models using the received global prediction. At each communication round $t (t\in\{1,2,...,T\})$, the training process is described as follows:
\begin{itemize}
    \item \textbf{Local training and prediction} Each client updates the parameters $\theta^k_t$ on its private data ${D^k}$ and takes predictions on the public data $D^0$. Then, the predictions are sent to the server.
        \begin{equation}
            \theta_t^k = \theta_{t-1}^k - \eta^k\nabla \mathcal {L}_{\text{CE}}(x_i^k,y_i^k;\theta_{t-1}^k),
            \label{eq:local-train}
        \end{equation}
        \begin{equation}
            z_t^k = f^k(x_i^0;\theta^k),
            \label{eq:local-pre}
        \end{equation}
    where $\eta^k$ is the learning rate in $k$th client, $\mathcal{L}_{\text{CE}}$ denotes the cross-entropy loss function between the predicted values and the ground truth, $z_t^k$ is the predicted logits on public data.
    \item \textbf{Server aggregation and distillation} Predictions from each client are aggregated by the central model for knowledge distillation. Each client acts as a teacher model while the server acts as a student model, learning from the teachers to train the student model.
        \begin{equation}
            z_t = \frac{1}{K} \sum_{k=1}^K z_t^k,
        \end{equation}
        \begin{equation}
            \theta_t = \theta_{t-1} - \eta\nabla\mathcal{L}_{\text{KL}}(z_t,f(x_i^0;\theta_{t-1})),
        \end{equation}
    where $z_t$ is aggregated predictions and $\mathcal{L}_{\text{KL}}(\cdot)$ denotes the Kullback-Leibler (KL) divergence loss function \cite{belov2011distributions}.
    \item \textbf{Local distillation} Each client performs distillation on the public data based on the global prediction $\widetilde{z}$, broadcast by the central model.
        \begin{equation}
            \theta_t^k = \theta_{t-1}^k - \eta^k\nabla \mathcal {L}_{\text{KL}}(\widetilde{z}, f^k(x_i^0;\theta_t^k)).
            \label{eq:local-distill}
        \end{equation}
\end{itemize}

\begin{figure}
    \centering
    \includegraphics[scale=0.09]{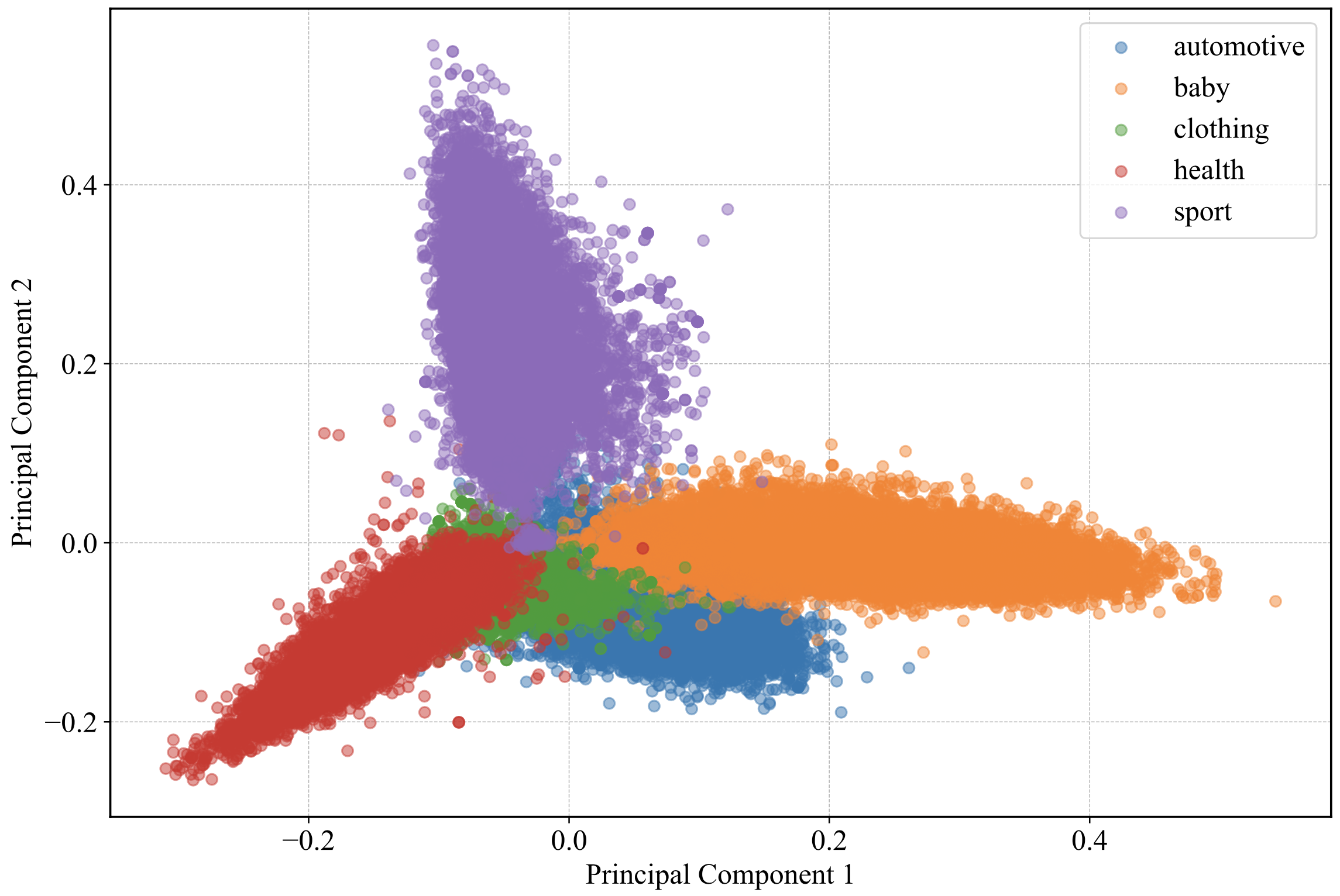}
    \caption{Diversity in language distributions across clients.}
    \label{fig:muti-domain}
\end{figure}

\section{Non-IID Benchmark in Natural Language Processing}
We introduce a more comprehensive definition of non-IID encompassing both diversity in label distributions and diversity in language domain across clients in natural language processing.

\subsection{Non-IID with Labels}
Following the previous work in federated distillation, we use Dirichlet distribution \cite{hsu2019measuring} to construct non-IID labels of the local data in clients. The Dirichlet distribution can effectively simulate the differences in label distribution among various clients and allows control over their diversity and skewness through the parameter vector $\alpha$. As $\alpha \to \infty$, the label distribution obtained by each client tends to become more evenly distributed, whereas $\alpha \to 0$, the label distribution is sharper.


\subsection{Non-IID on Languages}
To build non-IID on languages across clients, we diversify the vocabulary distributions over the data in clients. A domain-specific data is assigned to each client as its private data, obtaining non-IID inputs. From an open-source Amazon product review database\footnote{https://archive.org/details/amazon-reviews-1995-2013}, we randomly choose 5 different types of product reviews: automotive, baby, clothing, health and sport, and allocate these to 5 clients as local data, respectively. For each data, we randomly select 20\% samples without labels as public data and remain 80\% data as private data. Fig.~\ref{fig:muti-domain} visualizes the feature distributions of data in local clients, limiting the number of extracted features to 5K, where the 5 colors correspond to data from the different domains, and the extent of color overlap indicates the degree of similarity in the language domain across these data. 
\begin{algorithm}[h]
    \SetAlgoLined 
	\caption{Adaptive Federated Distillation}
        \label{alg:adafd}
	\KwIn{labeled private data $\{D^k\}_{k=1}^K$; unlabeled public data $D^0$; central model $\theta$; local models $\{\theta^k\}_{k=1}^K$; communication rounds $T$}
	\KwOut{central model $\theta$}
        \For{each communication round $t=1,2,...,T$}{
        \textbf{Local training and prediction}:
        
        \For{each client k in parallel}{
            training loss, $l_t^k,\theta_t^k \gets f^k(D^k;\theta_{t-1}^k)$ \;
            
            local predictions, $z_t^k \gets f^k(D^0;\theta_t^k)$ \;
            
        }
        \textbf{Server adaptive ensemble and distillation}:

        \For{each client k}{
            \eIf{LLMWC}{
            generate the weights by LLM \;
            }{
            $w_t^k \gets$ normalization$(l_t^k)$ via \textbf{Eq.}~(\ref{eq:ada-inverse-weight}) or \textbf{Eq.}~(\ref{eq:ada-exponent-weight}) \;
            
            $\widehat{z_t} \gets$ ensemble$(w_t^k,z_t^k)$ via \textbf{Eq.}~(\ref{eq:ada-ensemble}) \;
            } 
        }
        update the central model $\theta_{t-1}$ via \textbf{Eq.}~(\ref{eq:ada-distill}) \;

        global prediction $\widetilde{z_t} \gets f(D^0;\theta_t)$ \;
        
        \textbf{Local distillation}:

            \For{each client k in parallel}{
            update the local model $\theta_t^k$ by receiving $\widetilde{z_t}$ via \textbf{Eq.}~(\ref{eq:local-distill})\;
            }
        }
\end{algorithm}
Each point in the figure is obtained by reducing the TF-IDF matrix by Principal Components Analysis. 


\section{Adaptive Federated Distillation}
\begin{figure*}[ht]
    \centering
    \vspace{-1em}  
    \includegraphics[scale=0.6]{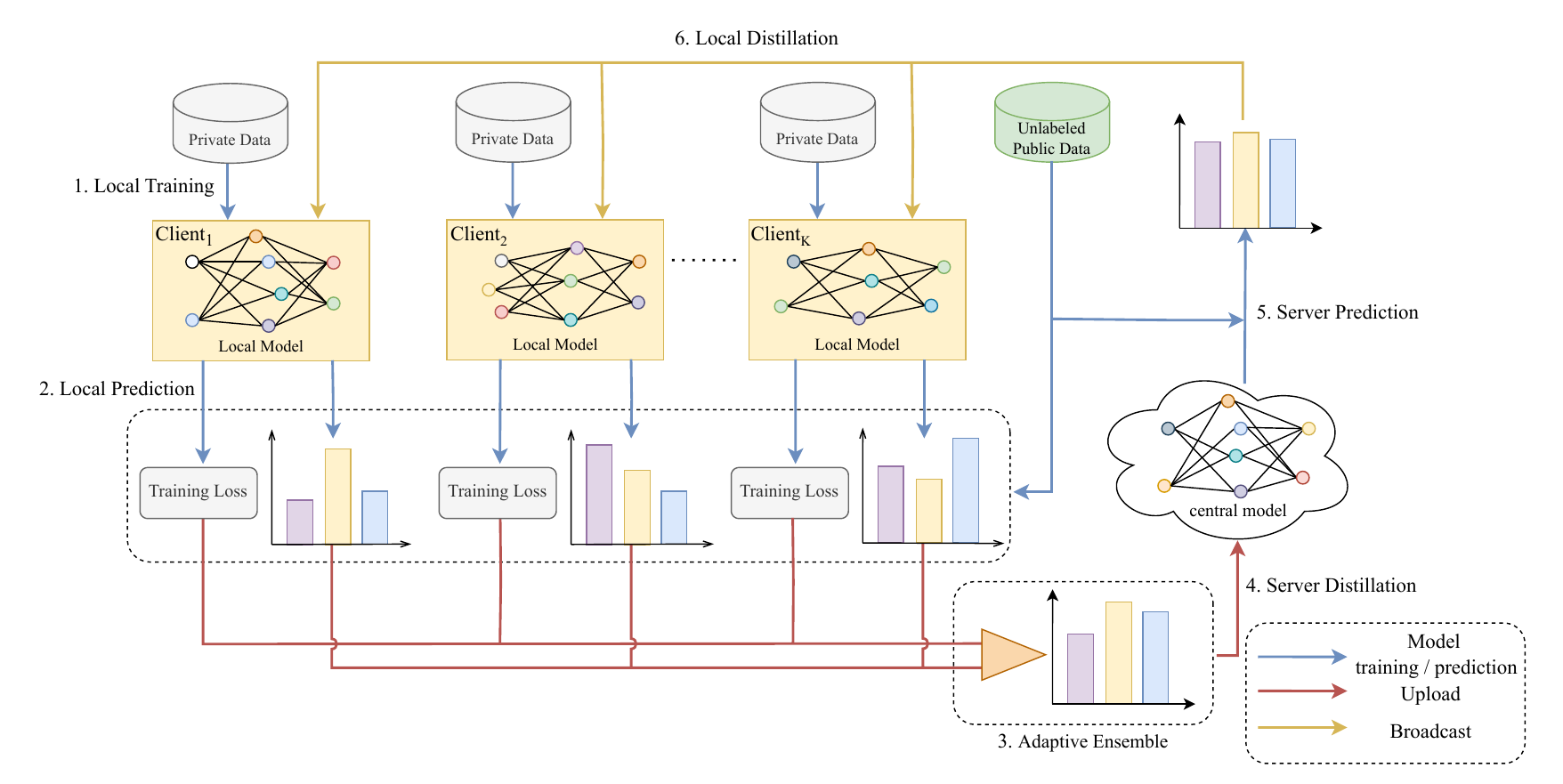}
    \vspace{-1em}  
    \caption{The framework of AdaFD.}
    \label{ada-fd}
\end{figure*}

We propose a novel framework of Adaptive Federate Distillation (AdaFD) to address the non-IID federated learning in natural language processing. The overall framework of AdaFD is shown in Fig.~\ref{ada-fd}. In a communication round $t~(t\in T)$, each local model is optimized on its labeled private data, and sends the predictions on the public data to a server. The predictions are aggregated in the server to train the central model through knowledge distillation with the proposed adaptive ensemble strategies. The global predictions given by the server are broadcast to all clients for local distillation. The detailed procedures are summarized in Algorithm~\ref{alg:adafd}.

\subsection{Adaptive Ensemble}
Predictions are aggregated according to the training losses in clients. Let $l^k$ be the training loss produced in the private data $D^k$ and $z^k = f(D^0;\theta^k)$ be the logits predicted in the unlabeled public data $D^0$ by the client $k$, respectively. 

We propose three adaptive ensemble strategies that compute the weights assigned to each client based on the training losses. These weights are accurately and dynamically adjusted according to the losses, where models that better align with the global data distribution are assigned greater weights, thereby enhancing the performance of the central model. The three proposed strategies are described as follows:

\begin{itemize}

\item \textbf{Reciprocal Normalized Weight Calculation (RNWC)} We obtain the weights by calculating the minimum training loss $l_{\text{min}}^k$ of each model over multiple epochs of local training, using an inverse proportional function: 
    \begin{equation}
    w^k = \frac{1/l^k_{\text{min}}}{\left\| \mathbf{v} \right\|_1}, \quad \mathbf{v} = \left[ 1/l^1_{\text{min}}, \dots, 1/l^k_{\text{min}} \right],
    \label{eq:ada-inverse-weight}
    \end{equation}
The aggregated prediction is obtained as the sum of the weighted predictions from each client.

\item \textbf{Exponential Normalized Weight Calculation (ENWC)} Different from RNWC only in how the weights assigned to each client are calculated, ENWC adopts an exponential decay function:
    \begin{equation}
    w^k = \frac{e^{-\beta l^k_{\text{min}}}}{\left\| \mathbf{v} \right\|_1}, \quad \mathbf{v} = \left[ e^{-\beta l^1_{\text{min}}}, \dots, e^{-\beta l^K_{\text{min}}} \right],
    \label{eq:ada-exponent-weight}
    \end{equation}

where hyperparameter $\beta$ affects the smoothness of the weight distribution and is used to adjust the weight allocation in the adaptive ensemble strategy of the model.\\

The weights calculated by the above two strategies are used to aggregate the logits predicted by the clients, resulting in adaptive ensemble logits:
    \begin{equation}
    \widehat{z} = \sum_{k=1}^K w^k z^k.
    \label{eq:ada-ensemble}
    \end{equation}

\item \textbf{Large Language Models Weight Calculation (LLMWC)}

Due to the success of LLM in NLP, we use to generate weights assigned to each client. To ensure more accurate analysis and computation, prompts are designed based on the training losses, F1 score, predicted logits, and the corresponding entropy of logits obtained during each training epoch. These prompts guide LLM to produce the weights, as detailed in the Appendix \ref{appendix:prompt}. 
\end{itemize}

\subsection{Adaptive Distillation}
The central model takes predictions $\widetilde{z} = f(D^0;\theta)$ on the unlabeled public data $D^0$ and is updated using the adaptive ensemble logits $\widehat{z}$ through adaptive distillation in a round $t$,

\begin{equation}
    \theta_t = \theta_{t-1} - \eta\nabla\mathcal{L} (\sigma(\widetilde{z}),\sigma(\widehat{z})).
    \label{eq:ada-distill}
\end{equation}
Traditional federated distillation usually uses KL divergence with a temperature parameter $\tau$ to distill the central model. To enhance the generalization and learnability of the central model, we consider the case where $\tau \to \infty$, under which the KL divergence can be approximated by the L2 loss.


\begin{equation}
    \mathcal{L} = {\left\| \widetilde{z} - \widehat{z} \right\|^2_2.}
    \label{eq:mse}
\end{equation}

\begin{table*}
    \caption{The experimental results on the client-specific test sets indicate that AdaFD achieves statistically significant improvements over baseline methods ($p < 0.001$), demonstrating the robustness and effectiveness of our method.}
    \resizebox{\textwidth}{!}{
    \begin{tabular}{c|c|ccccc||c|ccccc}
     \toprule
     & \multicolumn{6}{c||}{original scale} & \multicolumn{6}{c}{downsampled scale} \\
     Model & \textbf{\makecell{global \\ \small{(296K)}}} & \textbf{\makecell{auto. \\ \small{(33K)}}} & \textbf{\makecell{baby \\ \small(61K)}} & \textbf{\makecell{clot. \\ \small{(101K)}}} & \textbf{\makecell{heal. \\ \small{(34K)}}} & \textbf{\makecell{sport \\ \small{(67K)}}} & \textbf{\makecell{global \\ \small{(160K)}}} & \textbf{\makecell{auto. \\ \small(32K)}} & \textbf{\makecell{baby \\ \small{(32K)}}} & \textbf{\makecell{clot. \\ \small(32K)}} & \textbf{\makecell{heal. \\ \small{(32K)}}} & \textbf{\makecell{sport \\ \small{(32K)}}} \\
    \midrule
    \multicolumn{13}{c}{homogeneous} \\
    \midrule
    Centralized & 86.4 & 79.1 & 84.4 & 94.4 & 82.6 & 82.3 & 82.6 & 78.2 & 84.1 & 91.2 & 81.7 & 78.5  \\
    \midrule
    FedAvg & 79.1 & 77.8 & 81.5 & 80.5 & \textbf{77.1} & \textbf{79.7} & 76.2 & 76.1 & 80.2 & 76.6 & 76.6 & 76.8  \\
    DS-FL & 79.1 & 77.1 & 81.2 & 81.5 & 76.5 & 79.1 & 75.9 & 75.6 & 80.1 & 76.5 & 75.5 & 76.9  \\
    MHAT & 77.4 & 76.0 & 81.3 & 79.2 & 73.0 & 78.0 & 75.2 & 76.4 & 80.2 & 72.0 & 76.4 & 76.7  \\
    FedKD & 80.8 & 77.0 & \textbf{82.1} & 88.1 & 72.7 & 77.9 & 76.4 & 75.3 & 80.2 & 77.4 & \textbf{77.4} & \textbf{77.1}  \\
    \midrule
    AdaFD w/ LLMWC & 81.9 & 78.1 & \textbf{82.1} & 88.6 & 75.4 & 79.6 & 77.7 & \textbf{77.0} & 80.8 & 83.3 & 74.7 & 76.9  \\
    ~~~~~~~w/ RNWC & 81.4 & 76.4 & 79.7 & \textbf{91.0} & 73.7 & 78.2 & \textbf{78.0} & 76.0 & 80.4 & \textbf{86.0} & 75.5 & 76.5  \\
    ~~~~~~~w/ ENWC & \textbf{82.0} & \textbf{78.3} & 81.5 & 89.3 & 74.7 & \textbf{79.7} & \textbf{78.0} & 77.0 & \textbf{81.2} & 84.3 & 74.9 & 76.9  \\
    \midrule
    \midrule
    \multicolumn{13}{c}{heterogeneous} \\
    \midrule
    Centralized & 88.2 & 83.2 & 86.7 & 94.5 & 85.3 & 84.3 & 85.5 & 82.6 & 86.4 & 91.4 & 84.8 & 82.6  \\
    \midrule
    FedAvg & 81.3 & 80.6 & 85.0 & 81.1 & 79.3 & \textbf{82.7} & 80.3 & 80.9 & 84.8 & 80.6 & 79.1 & 80.2  \\
    DS-FL & 80.7 & 80.4 & 84.1 & 79.9 & \textbf{81.1} & 82.1 & 79.9 & 80.3 & 84.7 & 77.8 & \textbf{80.6} & 80.2  \\
    MHAT & 81.1 & 80.5 & 85.4 & 82.3 & 76.5 & 81.4 & 79.0 & 79.8 & 84.2 & 77.3 & 79.0 & 79.4  \\
    FedKD & 83.5 & 80.5 & 85.3 & 87.4 & 78.4 & 82.6 & 80.3 & 81.2 & 84.8 & 79.7 & 79.5 & 80.3  \\
    \midrule
   AdaFD w/ LLMWC & 83.4 & \textbf{81.5} & \textbf{85.6} & 86.9 & 78.2 & 82.3 & 80.7 & \textbf{81.3} & 84.1 & 84.4 & 77.5 & 80.4  \\
    ~~~~~~~w/ RNWC & \textbf{84.6} & 80.4 & 85.7 & 89.9 & 79.2 & 82.5 & \textbf{81.2} & 80.9 & 84.9 & 83.7 & 79.8 & 80.2  \\
    ~~~~~~~w/ ENWC & \textbf{84.6} & 81.0 & 85.6 & \textbf{90.1} & 78.2 & 81.7 & \textbf{81.2} & 80.7 & \textbf{85.0} & \textbf{84.7} & 78.7 & \textbf{80.5}  \\
    \bottomrule
    \end{tabular}}
    \label{tab:client-specific-results}
\end{table*}

\section{Experiments}
The experiments are carried out in the sentiment classification task in both homogeneous and heterogeneous settings to demonstrate the performance of our proposed models.

\subsection{Settings}
We perform experiments on the text sentiment classification task using the non-IID data introduced in the previous section with $\alpha = 1$.
We split the private data into training, development and test sets with an 8:1:1 ratio.
We investigate our proposed framework in homogeneous settings and heterogeneous settings.

\begin{itemize}
    \item \textbf{Homogeneous} The clients and the server have the same backbone of pre-trained language model. \texttt{BERT-base-cased} model is adopted in our experiments.
    \item \textbf{Heterogeneous} We configure the local models in the 5 clients with different pre-trained language models, namely \texttt{BERT-base-cased}, \texttt{BERT-large-cased}, \texttt{RoBERTa-base}, \texttt{RoBERTa-large}, and \texttt{XLNet-large-cased} model, respectively, while using \texttt{RoBERTa-large} as the central model for the server.\footnote{All the pre-trained language model cards are sourced from \url{https://huggingface.co/models}.}
\end{itemize}

\subsection{Training and Inference Details}
The models are trained for 5 communication rounds and 3 epochs of local training. The initial learning rate is 2e-5. We use AdamW as optimizer, set the maximum sentence length to 128, and use a batch size of 32. Meanwhile, the number of clients is fixed at 5, corresponding to the number of domain, and the adaptive ensemble hyperparameter $\beta$ and temperature $\tau$ is set to 5 and 1, respectively. We use GPT-4 as the LLM.

\subsection{Baselines}
We reimplement existing federated distillation frameworks that are designed for both homogeneous and heterogeneous settings as our baselines for comparison with our proposed method. These methods are described as follows:
\begin{itemize}
    \item \textbf{Centralized} The model employs centralized training, conducted on a global training, development, and test set composed of private data from all clients. 
    \item \textbf{FedAvg} The model aggregates predictions from each client on the public data $D^0$ by assigning equal weights $\frac{1}{K}$ to clients, and the resulting ensemble predictions are used to optimize the central model by KL divergence.
    \item \textbf{DS-FL} The model uses the size of the private data $D^k$ as weights and entropy reduction aggregation of logits strategy to aggregate predictions from the clients \cite{itahara2021distillation}.
    \item \textbf{MHAT} The model uses the size of the private data $D^k$ as weights to aggregate predictions from the clients, and the resulting ensemble predictions are used to optimize the central model by Cross-Entropy \cite{hu2021mhat}.
    \item \textbf{FedKD} The model uses the size of the private data $D^k$ as weights to aggregate predictions from the clients. The central model only performs a one-shot distillation step on the ensemble predictions using KL divergence \cite{gong2022preserving}.
\end{itemize}

\subsection{Results}
\subsubsection{Results on Client-specific Test Sets}
We evaluate the proposed model on the client-specific test sets, which are composed of the domains corresponding to each client. TABLE \ref{tab:client-specific-results} shows the F1 scores on the client-specific test sets under homogeneous and heterogeneous settings, including results on the global distribution (global) and on the local distributions, i.e., automotive (auto.), baby, clothing (clot.), health (heal.) and sport. Although our models have marginally lower performances on test data in certain local distributions, we obtain better global model. One reason is that the baselines can obtain the local optimization in these distributions instead of global one, resulting in the lack of the collaboration with other distributions to achieve the global optimized models.
In both homogeneous and heterogeneous settings, our models achieve the best results. AdaFD w/ ENWC and AdaFD w/ RNWC not only outperform AdaFD w/ LLMWC, but eliminate LLM computation response time and LLM API invocation fees. We take the AdaFD w/ ENWC as our final model.

\begin{table*}
    \caption{The experimental results on the global test set indicate that AdaFD achieves statistically significant improvements over baseline methods ($p < 0.001$), demonstrating the robustness and effectiveness of our method.}
    \resizebox{\textwidth}{!}{
    \begin{tabular}{c|c|ccccc||c|ccccc}
     \toprule
     & \multicolumn{6}{c||}{original scale} & \multicolumn{6}{c}{downsampled scale} \\
     Model & \textbf{\makecell{global \\ \small{(296K)}}} & \textbf{\makecell{auto. \\ \small{(33K)}}} & \textbf{\makecell{baby \\ \small(61K)}} & \textbf{\makecell{clot. \\ \small{(101K)}}} & \textbf{\makecell{heal. \\ \small{(34K)}}} & \textbf{\makecell{sport \\ \small{(67K)}}} & \textbf{\makecell{global \\ \small{(160K)}}} & \textbf{\makecell{auto. \\ \small(32K)}} & \textbf{\makecell{baby \\ \small{(32K)}}} & \textbf{\makecell{clot. \\ \small(32K)}} & \textbf{\makecell{heal. \\ \small{(32K)}}} & \textbf{\makecell{sport \\ \small{(32K)}}} \\
    \midrule
    \multicolumn{13}{c}{homogeneous} \\
    \midrule
    Centralized & 76.7 & 79.8 & 84.9 & 93.8 & 83.3 & 82.6 & 73.8 & 78.9 & 84.0 & 91.3 & 82.2 & 79.8  \\
    \midrule
    FedAvg & 77.6 & 76.9 & 81.6 & 82.0 & 75.6 & \textbf{79.4} & 75.7 & 76.6 & 80.6 & 77.9 & \textbf{76.7} & 77.0  \\
    DS-FL & 76.4 & 76.7 & 81.2 & 81.3 & \textbf{76.6} & 79.1 & 76.0 & 76.2 & 80.7 & 79.1 & 73.9 & 75.9  \\
    MHAT & 76.6 & 75.8 & 80.6 & 82.0 & 71.7 & 77.0 & 75.1 & 74.9 & 80.8 & 74.4 & 75.0 & 76.3  \\
    FedKD & 76.8 & 76.1 & 80.6 & 88.3 & 74.3 & 78.7 & 75.8 & 76.2 & 81.2 & 81.2 & 72.6 & 75.7  \\
    \midrule
    AdaFD w/ LLMWC & 77.4 & \textbf{78.0} & \textbf{82.4} & 88.9 & 73.3 & 78.6 & 76.9 & \textbf{77.3} & 81.6 & 83.6 & 73.5 & 76.8  \\
    ~~~~~~~w/ RNWC & 77.2 & 76.4 & 81.1 & \textbf{90.5} & 73.1 & 78.7 & 75.2 & 75.6 & 80.6 & \textbf{87.4} & 71.9 & 75.5  \\
    ~~~~~~~w/ ENWC & \textbf{78.0} & 77.8 & 82.0 & 89.1 & 74.6 & 79.2 & \textbf{77.4} & 76.6 & \textbf{81.7} & 83.7 & 75.7 & \textbf{77.2}  \\
    \midrule
    \midrule
    \multicolumn{13}{c}{heterogeneous} \\
    \midrule
    Centralized & 79.5 & 83.2 & 86.7 & 94.5 & 85.3 & 84.3 & 82.2 & 82.6 & 86.4 & 91.4 & 84.8 & 82.6  \\
    \midrule
    FedAvg & 80.7 & 81.0 & 85.3 & 83.0 & 79.3 & 82.7 & 80.3 & 79.9 & \textbf{84.3} & 78.9 & 81.0 & 79.6  \\
    DS-FL & 81.6 & 80.2 & 84.6 & 80.3 & \textbf{81.0} & 82.7 & 77.4 & 79.8 & 83.1 & 78.3 & \textbf{81.3} & 79.2  \\
    MHAT & 81.7 & 79.6 & 85.0 & 82.7 & 75.0 & 80.5 & \textbf{81.1} & 79.6 & 83.4 & 72.9 & 79.8 & 78.5  \\
    FedKD & 82.0 & 80.7 & 84.9 & 89.4 & 75.2 & 81.0 & 81.0 & 79.9 & \textbf{84.3} & 77.7 & 80.6 & 79.6  \\
    \midrule
   AdaFD w/ LLMWC & 80.1 & 80.6 & 84.4 & 88.9 & 78.9 & 82.2 & 79.1 & 80.0 & 83.5 & \textbf{84.7} & 78.4 & 78.8  \\
    ~~~~~~~w/ RNWC & 82.1 & 80.7 & 85.1 & \textbf{90.5} & 74.7 & 81.2 & 81.0 & 80.7 & \textbf{84.3} & 84.1 & 78.0 & 78.9  \\
    ~~~~~~~w/ ENWC & \textbf{82.5} & \textbf{81.2} & \textbf{85.8} & 89.2 & 79.4 & \textbf{82.9} & 79.9 & \textbf{80.9} & 84.0 & 84.3 & 80.3 & \textbf{80.1}  \\
    \bottomrule
    \end{tabular}}
    \label{tab:global-results}
\end{table*}

\begin{figure}[htbp]
    \centering{\includegraphics[scale=0.26]{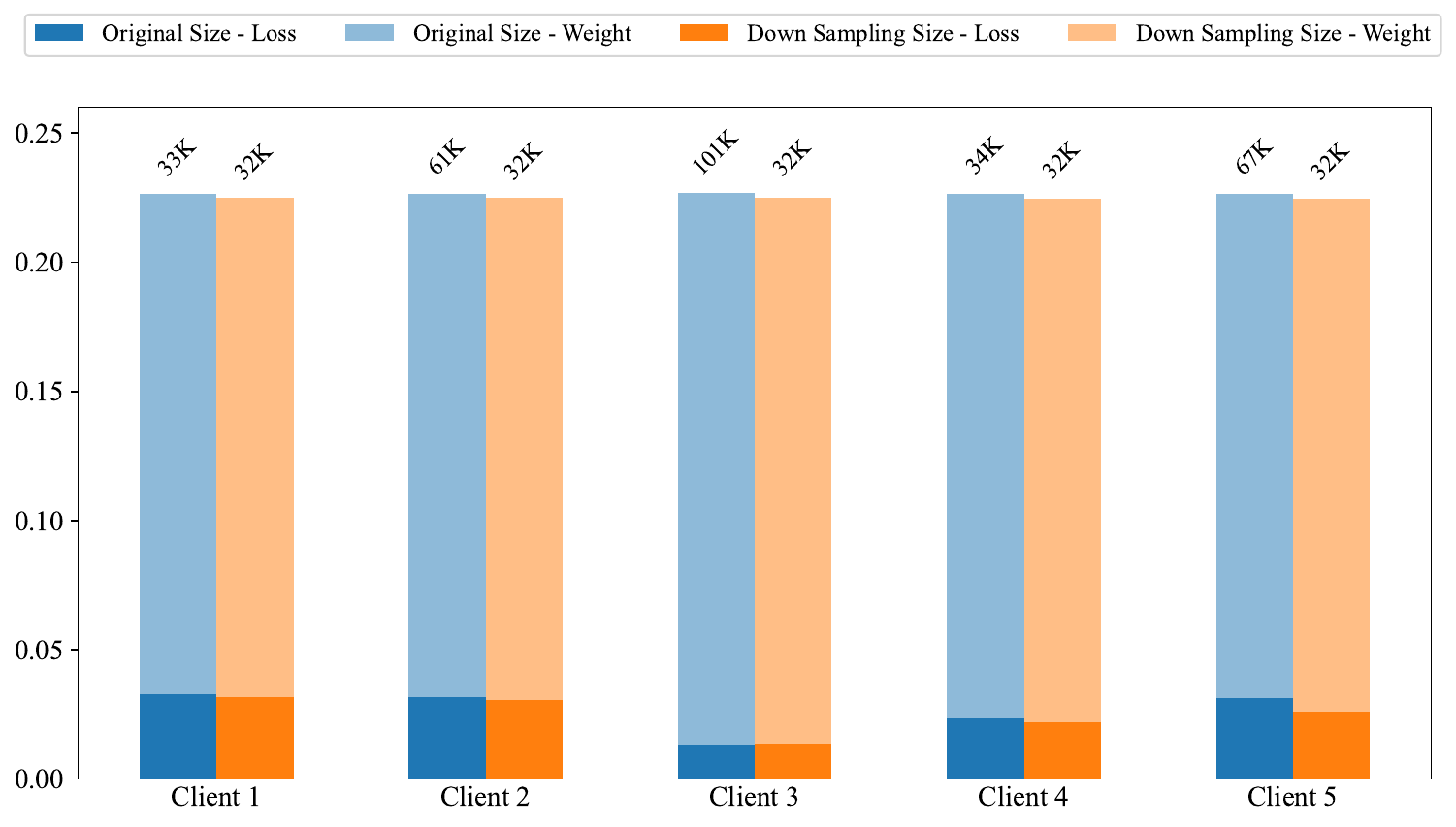}}
    \\
    \includegraphics[scale=0.26]{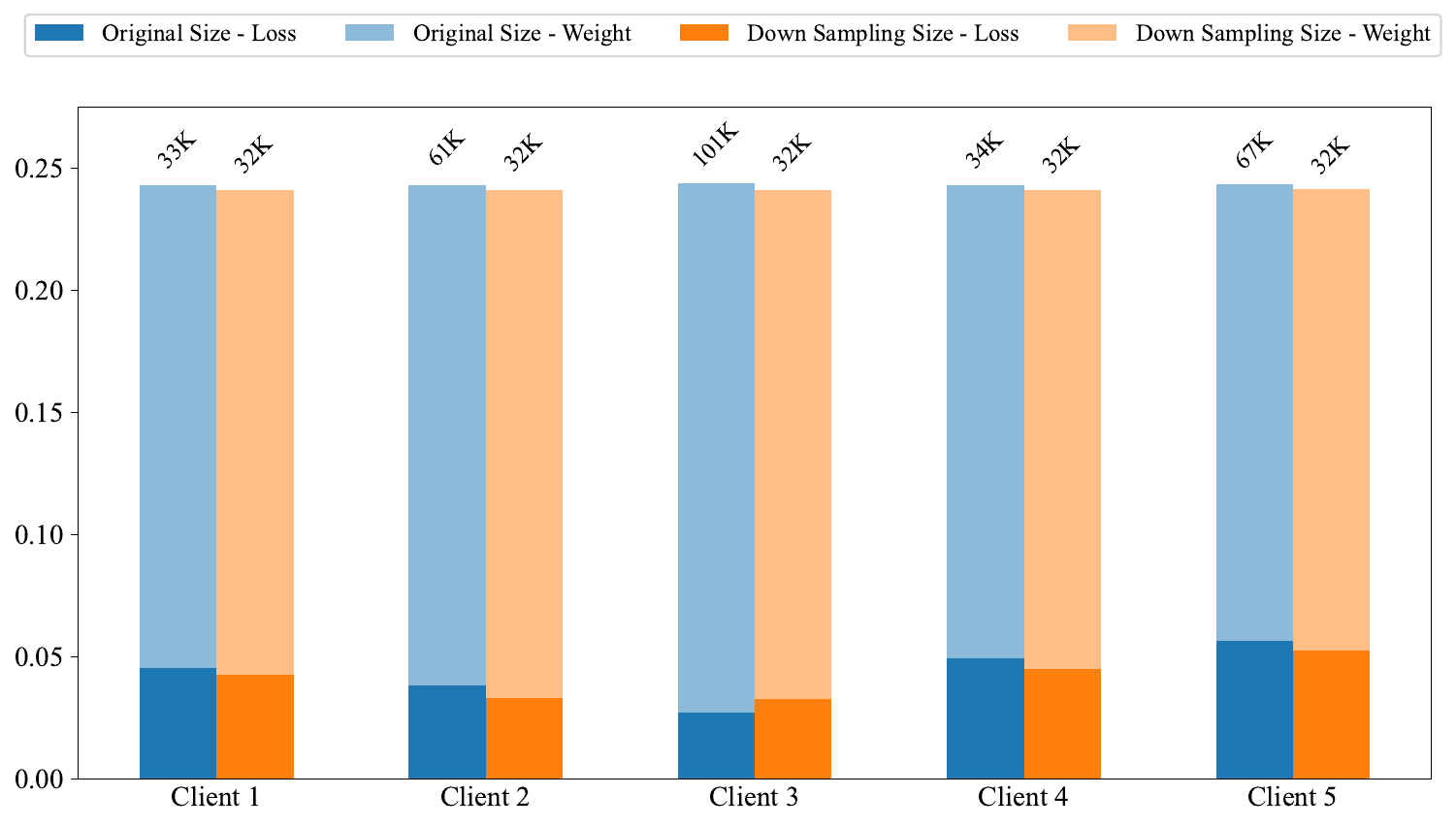}
    \caption{The losses and weights of varying data sizes on the central model in both homogeneous (upper) and heterogeneous (bottom) settings.}
    \label{fig:train_weight}
\end{figure}

\begin{itemize}
    \item \textbf{Homogeneous} 
    Centralized achieves the best results on both the global and individual data by sharing training data without preserving the privacy. AdaFD w/ ENWC achieves the best performance on non-IID global data in both original-scale and downsampling-scale data compared to other decentralized models. 
    \item \textbf{Heterogeneous} AdaFD w/ ENWC outperforms other decentralized models on the non-IID global data in heterogeneous settings.
\end{itemize}

\subsubsection{Results on Global Test Set}
We further evaluate the proposed model on the global test set, which is constructed from all domains of the open-source Amazon product review database. The global test set is sampled with $\alpha=0.8$ and its size is consistent with the client-specific test sets. TABLE \ref{tab:global-results} shows the F1 scores on the global test set under homogeneous and heterogeneous settings. Although the improvement is somewhat diminished in the heterogeneous downsampling-scale, AdaFD w/ ENWC still obtains the best results across both homogeneous and heterogeneous settings, further confirming the effectiveness and generalization capability of the proposed approach.

\begin{itemize}
    \item \textbf{Homogeneous} 
     AdaFD w/ ENWC achieves the best performance on non-IID global data in both original-scale and downsampling-scale data compared to centralized and other decentralized models.
    \item \textbf{Heterogeneous}
    Although the improvement is somewhat diminished in downsampling-scale, AdaFD w/ ENWC still outperforms other decentralized models on the non-IID global data in original-scale.
\end{itemize}

\section{Discussion and Analysis}
We take the further analysis on the performances of our final model in effect of data scale, effect of communication rounds, effect of adaptive ensemble, effect of hyperparameter and training losses.

\begin{figure*}[!htp]
    \centering
    \includegraphics[width=\textwidth]{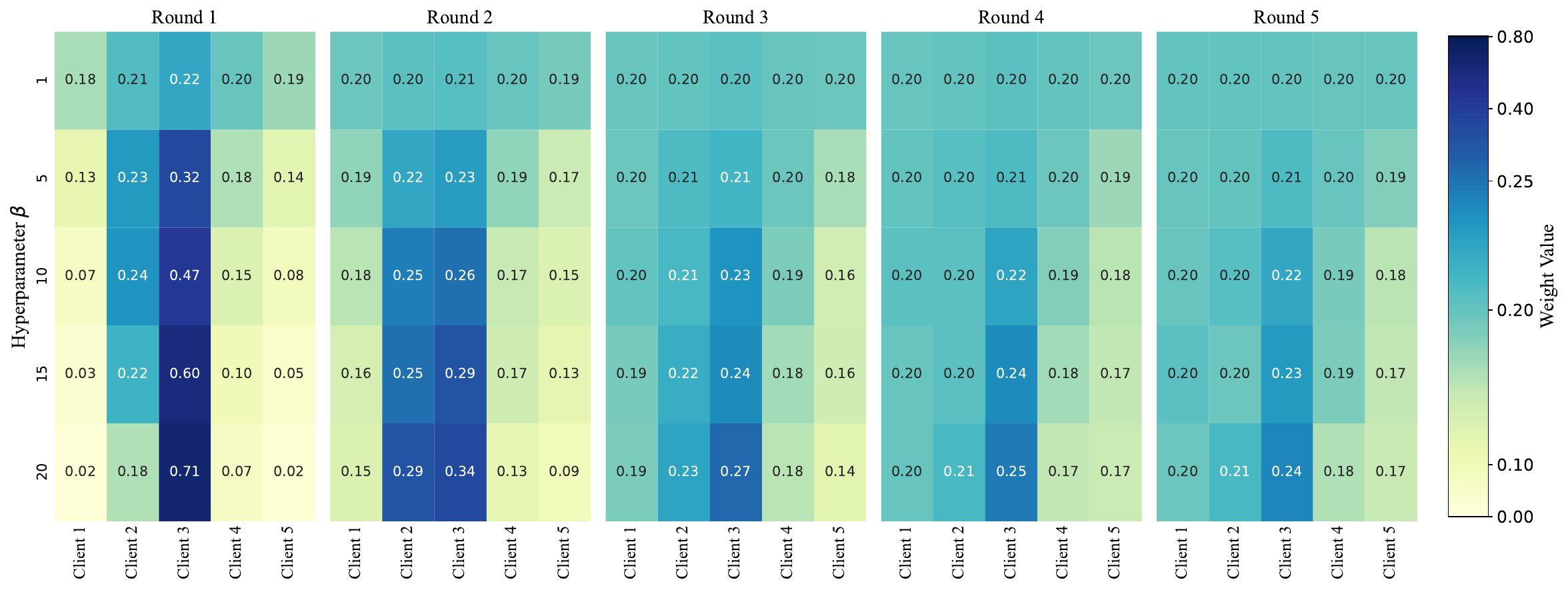}
    \caption{The weights vary with changes in the hyperparameter $\beta$ and across different rounds of federated distillation. Deeper color denotes a higher weight assigned to the corresponding client. }
    \label{fig:weights}
\end{figure*}

\begin{figure}[htbp]
\centerline{\includegraphics[scale=0.28]{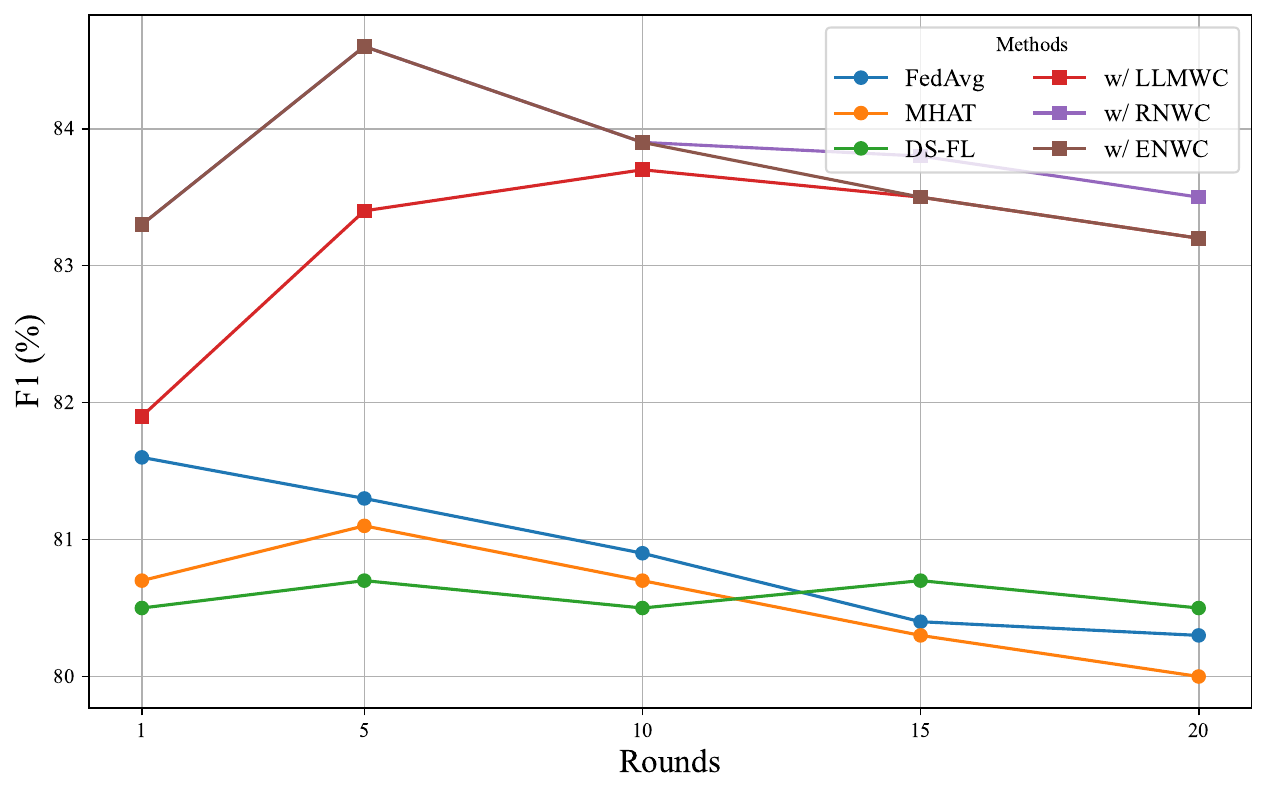}}
\caption{The F1 scores of various methods vary with the number of rounds}
\label{fig:multi_round_methods_cpm}
\end{figure}

\subsection{Effect of Data Scale}
To assess whether our models are dominated by local clients with larger scales of training data, we downsample 32K instances from each data according to label distribution, ensuring that the sampled data maintains their non-IID features. We conduct both homogeneous and heterogeneous classification experiments on these subsets. 

Fig.~\ref{fig:train_weight} shows the losses and the weights for each client under homogeneous and heterogeneous settings, respectively. Despite limiting the training data to 32K samples, the scale of the losses remained consistent, and the central model demonstrates better generalization on larger data, and the influence of data size on the effectiveness of the adaptive ensemble is minimal.

\subsection{Effect of Communication Rounds}
To investigate the convergence of the central model, the communication rounds are set to 1, 5, 10, 15, 20\footnote{we do not draw the changes of FedKD because it solely performs a one-shot distillation.}. As shown in Fig.~\ref{fig:multi_round_methods_cpm}, the F1 scores of the models generally improve with an increasing number of communication rounds, but after reaching 5 rounds, they begin to decline gradually. Most of the models achieve the highest F1 score at 5 communication rounds, where AdaFD exhibits superior performance and robustness.

\begin{figure}[htbp]
    \centering
    \begin{minipage}[b]{0.47\linewidth} 
        \centering
        \includegraphics[width=\linewidth]{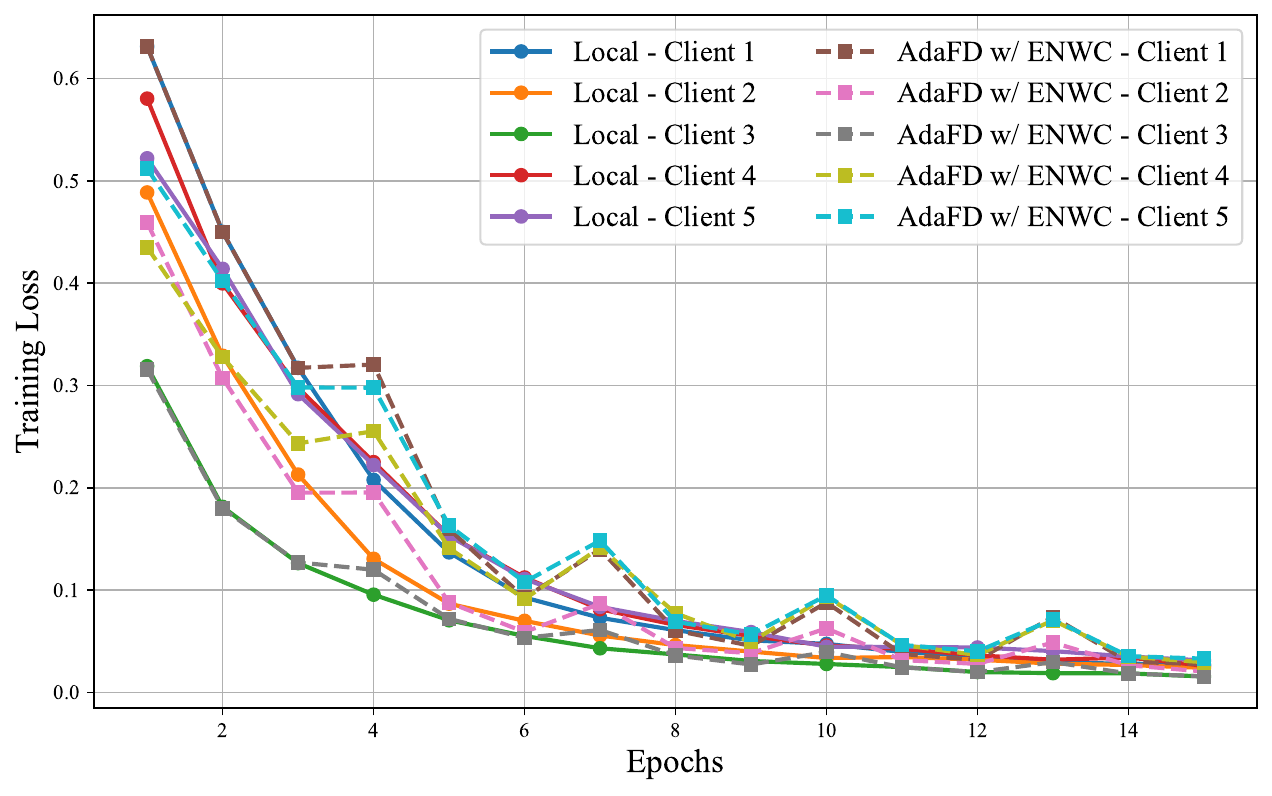}
    \end{minipage}
    \hspace{0.02\linewidth} 
    \begin{minipage}[b]{0.47\linewidth}
        \centering
        \includegraphics[width=\linewidth]{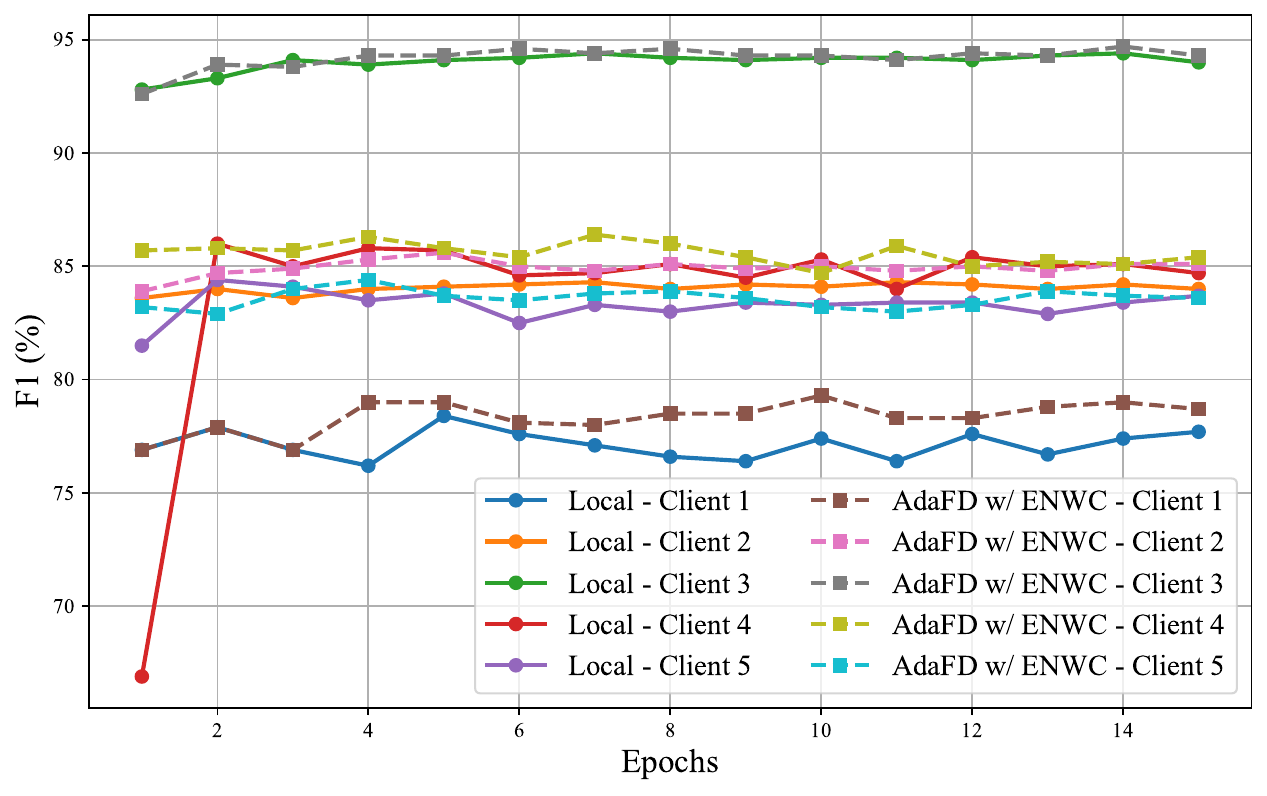} 
    \end{minipage}
    \caption{The training losses (left) and F1 scores (right) of Local and AdaFD vary with number of training epochs.}
    \label{fig:combined_loss_f1}
\end{figure}

\subsection{Effect of Adaptive Ensemble}
As shown in Fig.~\ref{fig:weights}, client 3 is initially assigned the highest weight during the early stages of communication with the server, but its weight decreases after several communication rounds. This is likely due to the fact that predictions from client 3 have the highest confidence level at the beginning. By the end of the training process, the weights converge to equal values across all clients, indicating that the model has achieved global optimization rather than being influenced by specific local distributions. We believe that the performance of the models can be further improved if the weights can be adjusted in finer grains, which is left to our future work.

\subsection{Effect of Hyperparameter}
The hyperparameter $\beta$ is a crucial variable in adaptive ensemble. To investigate the impact of $\beta$ on the performance of the central model, we set $\beta$ to 1, 5, 10, 15, and 20, and conduct experiments under heterogeneous settings and with original data scale.
As shown in Fig.~\ref{fig:weights}, we can observe that as $\beta$ increases, the weight distribution becomes sharper, meaning that the weight assigned to a particular client becomes larger compared to others. In contrast, as $\beta$ decreases, the weight distribution becomes more balanced, with weights being more evenly allocated between clients. 

With $\beta = 5$, the central model achieves its highest F1 score of 84.6, suggesting that clients with more accurate predictions should be assigned a larger weight. However, the performance of the central model declines if $\beta = 20$, because most of the weights are concentrated on a single client. This indicates that the domination by one client harms the global performance. We believe that the search space could be expanded if $\beta$ is learnable and fine-tuned during federated distillation, leading to a better performance, which is left to our future work.


\subsection{Training Losses}

Training losses are used for computing weights in adaptive ensemble, as in addition to the training losses at model initialization, they can also capture the fitting quality of the global data distribution, rather than being limited to the local private data. This weighting mechanism effectively reflects the influence of private data on the global data distribution, enabling the central model to better align with the global data distribution and achieve superior performance.

In Fig.~\ref{fig:combined_loss_f1}, Local denotes that each model is trained solely on its private data, whereas AdaFD initially uses private data for initialization but subsequently distills towards the global data distribution broadcast by the central model. The results clearly demonstrate that even though the training losses in AdaFD clients are higher than that in Local, the model performance of AdaFD is still slightly better. This strongly indicates that using training losses as a basis for weight calculation is both reasonable and effective.



\section{Conclusion}
In this paper, we provide an in-depth exploration of non-IID scenarios in natural language processing by addressing both language and label diversity, highlighting the challenges posed by heterogeneity in label and vocabulary distributions. We introduce a unified benchmarking framework that encompasses diverse data distributions across multiple domains, with the goal of advancing research on non-IID challenges in NLP. Additionally, we propose an adaptive federated distillation framework specifically tailored to address non-IID issues in both homogeneous and heterogeneous settings. Our experimental results demonstrate that our models outperform existing algorithms by capturing the diversity of clients in federated learning.

\section*{Acknowledgment}
We thank reviewers for their suggestions. This work is supported by Yunnan Fundamental Research Projects (grant NO.\ 202401CF070189)

\bibliographystyle{IEEEtran}
\bibliography{custom}

\begin{thebibliography}{10}
\providecommand{\url}[1]{#1}
\csname url@samestyle\endcsname
\providecommand{\newblock}{\relax}
\providecommand{\bibinfo}[2]{#2}
\providecommand{\BIBentrySTDinterwordspacing}{\spaceskip=0pt\relax}
\providecommand{\BIBentryALTinterwordstretchfactor}{4}
\providecommand{\BIBentryALTinterwordspacing}{\spaceskip=\fontdimen2\font plus
\BIBentryALTinterwordstretchfactor\fontdimen3\font minus \fontdimen4\font\relax}
\providecommand{\BIBforeignlanguage}[2]{{%
\expandafter\ifx\csname l@#1\endcsname\relax
\typeout{** WARNING: IEEEtran.bst: No hyphenation pattern has been}%
\typeout{** loaded for the language `#1'. Using the pattern for}%
\typeout{** the default language instead.}%
\else
\language=\csname l@#1\endcsname
\fi
#2}}
\providecommand{\BIBdecl}{\relax}
\BIBdecl

\bibitem{chik2013singapore}
W.~B. Chik, ``The singapore personal data protection act and an assessment of future trends in data privacy reform,'' \emph{Computer Law \& Security Review}, vol.~29, no.~5, pp. 554--575, 2013.

\bibitem{protection2018general}
F.~D. Protection, ``General data protection regulation,'' \emph{Intersoft Consulting, Accessed in October}, vol.~24, no.~1, 2018.

\bibitem{mcmahan2017communication}
B.~McMahan, E.~Moore, D.~Ramage, S.~Hampson, and B.~A. y~Arcas, ``Communication-efficient learning of deep networks from decentralized data,'' in \emph{Artificial intelligence and statistics}.\hskip 1em plus 0.5em minus 0.4em\relax PMLR, 2017, pp. 1273--1282.

\bibitem{tian2022fedbert}
Y.~Tian, Y.~Wan, L.~Lyu, D.~Yao, H.~Jin, and L.~Sun, ``Fedbert: When federated learning meets pre-training,'' \emph{ACM Transactions on Intelligent Systems and Technology (TIST)}, vol.~13, no.~4, pp. 1--26, 2022.

\bibitem{dong2022collaborating}
C.~Dong, Y.~Xie, B.~Ding, Y.~Shen, and Y.~Li, ``Collaborating heterogeneous natural language processing tasks via federated learning,'' \emph{arXiv preprint arXiv:2212.05789}, 2022.

\bibitem{zhang2022federated}
Z.~Zhang, X.~Hu, L.~Qu, Q.~Wang, and Z.~Xu, ``Federated model decomposition with private vocabulary for text classification,'' in \emph{Empirical Methods in Natural Language Processing 2022}.\hskip 1em plus 0.5em minus 0.4em\relax Association for Computational Linguistics (ACL), 2022, pp. 6413--6425.

\bibitem{lin2022fednlp}
B.~Y. Lin, C.~He, Z.~Ze, H.~Wang, Y.~Hua, C.~Dupuy, R.~Gupta, M.~Soltanolkotabi, X.~Ren, and S.~Avestimehr, ``Fednlp: Benchmarking federated learning methods for natural language processing tasks,'' in \emph{Findings of the Association for Computational Linguistics: NAACL 2022}, 2022, pp. 157--175.

\bibitem{abadeer2022flightner}
M.~Abadeer, W.~Shi, and J.-P. Corriveau, ``Flightner: A federated learning approach to lightweight named-entity recognition,'' in \emph{2022 IEEE International Conference on Trust, Security and Privacy in Computing and Communications (TrustCom)}.\hskip 1em plus 0.5em minus 0.4em\relax IEEE, 2022, pp. 687--694.

\bibitem{ma-etal-2023-fedid}
\BIBentryALTinterwordspacing
X.~Ma, J.~Liu, J.~Wang, and X.~Zhang, ``{F}ed{ID}: Federated interactive distillation for large-scale pretraining language models,'' in \emph{Proceedings of the 2023 Conference on Empirical Methods in Natural Language Processing}.\hskip 1em plus 0.5em minus 0.4em\relax Singapore: Association for Computational Linguistics, Dec. 2023, pp. 8566--8577. [Online]. Available: \url{https://aclanthology.org/2023.emnlp-main.529}
\BIBentrySTDinterwordspacing

\bibitem{jeong2018communication}
E.~Jeong, S.~Oh, H.~Kim, J.~Park, M.~Bennis, and S.-L. Kim, ``Communication-efficient on-device machine learning: Federated distillation and augmentation under non-iid private data,'' in \emph{Proceedings of the 32nd Conference on Neural Information Processing Systems}, 2018.

\bibitem{li2019fedmd}
D.~Li and J.~Wang, ``Fedmd: Heterogenous federated learning via model distillation,'' in \emph{Workshop on Federated Learning for Data Privacy and Confidentiality, NeurIPS 2019}, 2019.

\bibitem{chang2019cronus}
H.~Chang, V.~Shejwalkar, R.~Shokri, and A.~Houmansadr, ``Cronus: Robust and heterogeneous collaborative learning with black-box knowledge transfer,'' \emph{arXiv preprint arXiv:1912.11279}, 2019.

\bibitem{itahara2021distillation}
S.~Itahara, T.~Nishio, Y.~Koda, M.~Morikura, and K.~Yamamoto, ``Distillation-based semi-supervised federated learning for communication-efficient collaborative training with non-iid private data,'' \emph{IEEE Transactions on Mobile Computing}, vol.~22, no.~1, pp. 191--205, 2021.

\bibitem{hu2021mhat}
L.~Hu, H.~Yan, L.~Li, Z.~Pan, X.~Liu, and Z.~Zhang, ``Mhat: An efficient model-heterogenous aggregation training scheme for federated learning,'' \emph{Information Sciences}, vol. 560, pp. 493--503, 2021.

\bibitem{gong2022preserving}
X.~Gong, A.~Sharma, S.~Karanam, Z.~Wu, T.~Chen, D.~Doermann, and A.~Innanje, ``Preserving privacy in federated learning with ensemble cross-domain knowledge distillation,'' in \emph{Proceedings of the AAAI Conference on Artificial Intelligence}, vol.~36, no.~11, 2022, pp. 11\,891--11\,899.

\bibitem{fishman2020speaks}
J.~A. Fishman, ``Who speaks what language to whom and when?'' in \emph{The bilingualism reader}.\hskip 1em plus 0.5em minus 0.4em\relax Routledge, 2020, pp. 55--70.

\bibitem{ge2020fedner}
S.~Ge, F.~Wu, C.~Wu, T.~Qi, Y.~Huang, and X.~Xie, ``Fedner: Privacy-preserving medical named entity recognition with federated learning,'' \emph{arXiv preprint arXiv:2003.09288}, 2020.

\bibitem{sui2020feded}
D.~Sui, Y.~Chen, J.~Zhao, Y.~Jia, Y.~Xie, and W.~Sun, ``Feded: Federated learning via ensemble distillation for medical relation extraction,'' in \emph{Proceedings of the 2020 conference on empirical methods in natural language processing (EMNLP)}, 2020, pp. 2118--2128.

\bibitem{wu2023faster}
X.~Wu, F.~Huang, Z.~Hu, and H.~Huang, ``Faster adaptive federated learning,'' in \emph{Proceedings of the AAAI conference on artificial intelligence}, vol.~37, no.~9, 2023, pp. 10\,379--10\,387.

\bibitem{huang2023rethinking}
W.~Huang, M.~Ye, Z.~Shi, H.~Li, and B.~Du, ``Rethinking federated learning with domain shift: A prototype view,'' in \emph{2023 IEEE/CVF Conference on Computer Vision and Pattern Recognition (CVPR)}.\hskip 1em plus 0.5em minus 0.4em\relax IEEE, 2023, pp. 16\,312--16\,322.

\bibitem{liu2024vertical}
Y.~Liu, Y.~Kang, T.~Zou, Y.~Pu, Y.~He, X.~Ye, Y.~Ouyang, Y.-Q. Zhang, and Q.~Yang, ``Vertical federated learning: Concepts, advances, and challenges,'' \emph{IEEE Transactions on Knowledge and Data Engineering}, pp. 3615--3634, 2024.

\bibitem{yazdinejad2024robust}
A.~Yazdinejad, A.~Dehghantanha, H.~Karimipour, G.~Srivastava, and R.~M. Parizi, ``A robust privacy-preserving federated learning model against model poisoning attacks,'' \emph{IEEE Transactions on Information Forensics and Security}, 2024.

\bibitem{liu2021federated}
M.~Liu, S.~Ho, M.~Wang, L.~Gao, Y.~Jin, and H.~Zhang, ``Federated learning meets natural language processing: A survey,'' \emph{arXiv preprint arXiv:2107.12603}, 2021.

\bibitem{carlini2021extracting}
N.~Carlini, F.~Tramer, E.~Wallace, M.~Jagielski, A.~Herbert-Voss, K.~Lee, A.~Roberts, T.~Brown, D.~Song, U.~Erlingsson \emph{et~al.}, ``Extracting training data from large language models,'' in \emph{30th USENIX Security Symposium (USENIX Security 21)}, 2021, pp. 2633--2650.

\bibitem{lin2020ensemble}
T.~Lin, L.~Kong, S.~U. Stich, and M.~Jaggi, ``Ensemble distillation for robust model fusion in federated learning,'' \emph{Advances in neural information processing systems}, vol.~33, pp. 2351--2363, 2020.

\bibitem{tan2022fedproto}
Y.~Tan, G.~Long, L.~Liu, T.~Zhou, Q.~Lu, J.~Jiang, and C.~Zhang, ``Fedproto: Federated prototype learning across heterogeneous clients,'' in \emph{Proceedings of the AAAI Conference on Artificial Intelligence}, vol.~36, no.~8, 2022, pp. 8432--8440.

\bibitem{zhang2023distill}
Y.~Zhang, W.~Zhang, L.~Pu, T.~Lin, and J.~Yan, ``To distill or not to distill: towards fast, accurate and communication efficient federated distillation learning,'' \emph{IEEE Internet of Things Journal}, 2023.

\bibitem{li2023feddkd}
X.~Li, B.~Chen, and W.~Lu, ``Feddkd: Federated learning with decentralized knowledge distillation,'' \emph{Applied Intelligence}, vol.~53, no.~15, pp. 18\,547--18\,563, 2023.

\bibitem{wang2023dafkd}
H.~Wang, Y.~Li, W.~Xu, R.~Li, Y.~Zhan, and Z.~Zeng, ``Dafkd: Domain-aware federated knowledge distillation,'' in \emph{Proceedings of the IEEE/CVF conference on computer vision and pattern recognition}, 2023, pp. 20\,412--20\,421.

\bibitem{yang2024fedfed}
Z.~Yang, Y.~Zhang, Y.~Zheng, X.~Tian, H.~Peng, T.~Liu, and B.~Han, ``Fedfed: Feature distillation against data heterogeneity in federated learning,'' \emph{Advances in Neural Information Processing Systems}, vol.~36, 2024.

\bibitem{zhou2024federated}
X.~Zhou, W.~Huang, W.~Liang, Z.~Yan, J.~Ma, Y.~Pan, I.~Kevin, and K.~Wang, ``Federated distillation and blockchain empowered secure knowledge sharing for internet of medical things,'' \emph{Information Sciences}, vol. 662, p. 120217, 2024.

\bibitem{han2024fedal}
P.~Han, X.~Shi, and J.~Huang, ``Fedal: Black-box federated knowledge distillation enabled by adversarial learning,'' \emph{IEEE Journal on Selected Areas in Communications}, 2024.

\bibitem{kairouz2021advances}
P.~Kairouz, H.~B. McMahan, B.~Avent, A.~Bellet, M.~Bennis, A.~N. Bhagoji, K.~Bonawitz, Z.~Charles, G.~Cormode, R.~Cummings \emph{et~al.}, ``Advances and open problems in federated learning,'' \emph{Foundations and trends{\textregistered} in machine learning}, vol.~14, no. 1--2, pp. 1--210, 2021.

\bibitem{li2021survey}
Q.~Li, Z.~Wen, Z.~Wu, S.~Hu, N.~Wang, Y.~Li, X.~Liu, and B.~He, ``A survey on federated learning systems: Vision, hype and reality for data privacy and protection,'' \emph{IEEE Transactions on Knowledge and Data Engineering}, vol.~35, no.~4, pp. 3347--3366, 2021.

\bibitem{li2021model}
Q.~Li, B.~He, and D.~Song, ``Model-contrastive federated learning,'' in \emph{Proceedings of the IEEE/CVF conference on computer vision and pattern recognition}, 2021, pp. 10\,713--10\,722.

\bibitem{wang2021addressing}
L.~Wang, S.~Xu, X.~Wang, and Q.~Zhu, ``Addressing class imbalance in federated learning,'' in \emph{Proceedings of the AAAI Conference on Artificial Intelligence}, vol.~35, no.~11, 2021, pp. 10\,165--10\,173.

\bibitem{zhou2023efficient}
Z.~Zhou, S.~S. Azam, C.~G. Brinton, and D.~I. Inouye, ``Efficient federated domain translation,'' in \emph{The Eleventh International Conference on Learning Representations}, 2023.

\bibitem{azam2023federated}
S.~S. Azam, M.~Pelikan, V.~Feldman, K.~Talwar, J.~Silovsky, and T.~Likhomanenko, ``Federated learning for speech recognition: Revisiting current trends towards large-scale asr,'' in \emph{International Workshop on Federated Learning in the Age of Foundation Models in Conjunction with NeurIPS 2023}, 2023.

\bibitem{yang2023personalized}
L.~Yang, J.~Huang, W.~Lin, and J.~Cao, ``Personalized federated learning on non-iid data via group-based meta-learning,'' \emph{ACM Transactions on Knowledge Discovery from Data}, vol.~17, no.~4, pp. 1--20, 2023.

\bibitem{arafeh2023data}
M.~Arafeh, H.~Ould-Slimane, H.~Otrok, A.~Mourad, C.~Talhi, and E.~Damiani, ``Data independent warmup scheme for non-iid federated learning,'' \emph{Information Sciences}, vol. 623, pp. 342--360, 2023.

\bibitem{lu2024federated}
Z.~Lu, H.~Pan, Y.~Dai, X.~Si, and Y.~Zhang, ``Federated learning with non-iid data: A survey,'' \emph{IEEE Internet of Things Journal}, 2024.

\bibitem{li2024feature}
Z.~Li, Y.~Sun, J.~Shao, Y.~Mao, J.~H. Wang, and J.~Zhang, ``Feature matching data synthesis for non-iid federated learning,'' \emph{IEEE Transactions on Mobile Computing}, 2024.

\bibitem{belov2011distributions}
D.~I. Belov and R.~D. Armstrong, ``Distributions of the kullback--leibler divergence with applications,'' \emph{British Journal of Mathematical and Statistical Psychology}, vol.~64, no.~2, pp. 291--309, 2011.

\bibitem{hsu2019measuring}
T.-M.~H. Hsu, H.~Qi, and M.~Brown, ``Measuring the effects of non-identical data distribution for federated visual classification,'' \emph{arXiv preprint arXiv:1909.06335}, 2019.

\end{thebibliography}

\appendices
\section{Prompt Template}\label{appendix:prompt}
\begin{tcolorbox}[colback=gray!5!white,colframe=black!75!black]
You are an expert in federated learning and federated distillation. Analyze and compute the optimal weight allocation for all clients based on the following data in a federated distillation scenario. Return only the numerical values separated by commas, without any explanation or calculation process.

\textbf{Input:}
QUERY + \texttt{
Based on these factors, please compute and normalize the weight for each client to maximize the global model performance. Assign weights to each client and return the values separated by commas. Do not include any additional text or calculation details.}

\end{tcolorbox}

\noindent\textit{Note:} The input QUERY is dynamically constructed during model training. For each client $k$, we extract its local statistics—including training loss, F1 scores, logits, and logits entropy—and concatenate them to form the input prompt.

\end{document}